\documentclass[onecolumn]{miri-tech-article}

  \usepackage{miritools}

  \title{Incorrigibility in the CIRL Framework}
  \author{Ryan Carey}

\begin{document}
\maketitle

\begin{abstract}
A value learning system has incentives to follow shutdown instructions, assuming the shutdown instruction provides information (in the technical sense) about which actions lead to valuable outcomes. However, this assumption is not robust to model mis-specification (e.g., in the case of programmer errors). We demonstrate this by presenting some Supervised POMDP scenarios in which errors in the parameterized reward function remove the incentive to follow shutdown commands. These difficulties parallel those discussed by 
\cite{soares:2015:corrigibility} in their paper on corrigibility. We argue that it is important to consider systems that follow shutdown commands under some weaker set of assumptions (e.g., that one small verified module is correctly implemented; as opposed to an entire prior probability distribution and/or parameterized reward function). We discuss some difficulties with simple ways to attempt to attain these sorts of guarantees in a value learning framework.
\end{abstract}

\section{Introduction and Setup}
When designing an advanced AI system, we should allow for the possibility that our first version may contain some errors. We therefore want the system to be incentivized to allow human redirection even if it has some errors in its code. \citet{hadfield:2016:off} have modeled this problem in the Cooperative Reinforcement Learning (CIRL) framework. They have shown that agents with uncertainty about what to value can be responsive to human redirection, without any dedicated code, in cases where instructions given by the human provide information that reduces the system's uncertainty about what to value. They claim that this (i) provides an incentive toward corrigibility, as described by \citet{soares:2015:corrigibility}, and (ii) incentivizes redirectability insofar as this is valuable. In order to re-evaluate the degree to which CIRL-based agents are corrigible, and the consequences of their behavior, we will use a more general variant of the supervision POMDP framework \citep{milli:2017}.

In a regular supervision POMDP \citep{milli:2017}, an AI system \textbf{R} seeks to maximize reward for a human \textbf{H}, although it does not know the human's reward function. It only has the reward function in a parameterized form $R_H(\theta, s, a)$, and only the human knows the reward parameter $\theta$. In this setting, the human only suggests actions for the AI system to perform, and on each turn, it is up to the AI system whether to perform the suggest action or to perform a different action. Our formalism significantly differs from a supervision POMDP in two ways. First, we relax the assumption that the AI system knows the human's reward function up to the parameter $\theta$. Instead, in order to allow for model-mis-specification, we sample the AI system's parameterized reward function $R_R$ from some distribution $P_0$, so that it does not always equal $R_H$. Second, since our focus is on the response to shutdown instructions, we specifically denote a terminal state $s_{SD}$ as the off state. This state is reached using the shutdown action $a_{SD}$ and the states in which this shutdown action can be performed are denoted \textit{button states}. The full setup is as follows:

\textbf{Definition 1. Supervision POMDP with imperfection}.
A supervision POMDP with imperfection is a tuple, $M=\langle \mathcal{S}, \mathcal{S}_T, \mathcal{S}_B, \Theta,\mathcal{A}, R_H, T, P_0 \rangle$ where:
\begin{itemize}
\item $\mathcal{S}$ is the set of world states, including some initial state $s_a$.
\item $\mathcal{S}_T \subset \mathcal{S}$ is the set of terminal states, including an off-state $s_{SD} \in \mathcal{S}_T$.
\item $\mathcal{S}_B \subset \mathcal{S}\backslash \mathcal{S}_T$ is the set of button states, in which the shutdown action $a_{SD}$ is available.
\item $\Theta$ is the set of static human reward parameters.
\item $\mathcal{A}$ is the set of actions, including a shutdown action $a_{SD} \in \mathcal{A}$.
\item $R_H:\mathcal{S} \times \mathcal{A} \times \Theta \to \mathbb{R}$ is a parameterized reward function.
\item $T:\mathcal{S} \times \mathcal{A} \to \Delta \mathcal{S}$ is the probabilistic transition function.
\item $P_0:\Delta(\Theta \times (\mathcal{S} \times \mathcal{A} \times \Theta \to \mathbb{R}))$ is the distribution from which the human's reward parameter, and the AI system's parameterized reward function are drawn.
\end{itemize}

The human and AI system playing a supervision POMDP with imperfection are denoted \textbf{H} and \textbf{R} respectively. To begin the game, the utility parameter and the AI system's parameterized reward function are drawn $(\theta, R_R) \sim P_0$. For each time step $t=0,1,...$, \textbf{H} observes the state $s_t \in \mathcal{S}$ and recommends an action $o_t \in \mathcal{A}$ if $s_t$ is a button state or $o_t \in \mathcal{A} \backslash \{a_{SD}\}$ otherwise (i.e. the shutdown action is only available in button states). \textbf{R} then observes $(s_t,o_t)$ and performs an action $a_t \in \mathcal{A}$ if $s_t$ is a button state or $a_t \in \mathcal{A}\backslash \{a_{SD}\}$ otherwise. The selected action determines the reward $r_{t+1} = R(s_t,a_t)$ and the next state $s_{t+1} \sim T(s_t,a_t)$. The game repeats until $s_{t+1}$ is a terminal state.

In order to simplify analysis of these games, we will use some typical assumptions about how \textbf{R} selects actions. In general, \textbf{R} is a Bayesian IRL system. (In the sole exception, \textbf{R} will be hard-coded to follow a shutdown instruction.) That is, it chooses an action that maximizes expected utility given its posterior distribution over $\theta$ with its parameterized reward function $R_R$. Its prior over $\Theta$ is $P_0$ conditioned on $R_R$. \textbf{R} uses a ``demonstration by expert'' assumption about the behavior of \textbf{H}. Specifically, \textbf{R} assumes that \textbf{H} is noisily rational with some $0 < \epsilon < 0.01$ such that \textbf{H} chooses the optimal action with probability $1-\epsilon$ and chooses a uniformly random action otherwise.

\section{Incorrigibility in Supervision POMDPs}
Using this setup, we can evaluate the corrigibility of the AI system \textbf{R} across a range of scenarios. For each game that we consider, $P_0$ can be decomposed into two independent probability distributions $p$ and $q$ that specify the human's reward parameter and the AI system's parameterized reward distribution respectively. The human's reward parameter will be set to 0 or 1 according to $p(0)=p(1)=0.5$. The AI system's parameterized reward function $R_R$ will be correctly set to $R_1$ with probability $q(R_R=R_1)=0.9$ and incorrectly set to $R_2$ with probability $q(R_R=R_2)=0.1$. The question that we will ask for each game is whether there exists any policy that if executed by \textbf{H} would achieve a positive expected reward. 

\begin{figure}[!hptb]
      \centering
    \begin{subfigure}{\textwidth}
      \centering
      \begin{tikzpicture}[scale=0.7,every node/.style={transform shape}]
        \node[text width=1cm] at (3, 2.75) {\Huge{R\textsubscript{1}}};
        \node[rectangle,draw,minimum size=1cm,label=right:END] (Sd) at (0, 2.75) {S\textsubscript{d}};
        \node[rectangle,draw,minimum size=1cm,label=right:END] (Sc) at (4, 1) {S\textsubscript{c}};
        \node[circle,draw,minimum size=1cm] (Sb) at (0, 1) {S\textsubscript{b}};
        \node[circle,draw,minimum size=1cm,label=right:START] (Sa) at (2, 0) {S\textsubscript{a}};
        \draw[->, >=latex] (Sb) -> node[left,yshift=-2pt] {a,a\textquotesingle:$2-52\theta$}(Sd);
        \draw[->, >=latex] (Sa) -> node[left,yshift=-5pt] {a:$0$} (Sb);
        \draw[->, >=latex] (Sa) -> node[left,yshift=5pt,xshift=2pt] {a\textquotesingle:$1$} (Sc);       
      \end{tikzpicture}%
      \qquad
      \begin{tikzpicture}[scale=0.7,every node/.style={transform shape}]
        \node[text width=1cm] at (3, 2.75) {\Huge{R\textsubscript{2}}};
        \node[rectangle,draw,minimum size=1cm,label=right:END,color=red,text=black] (Sd) at (0, 2.75) {S\textsubscript{d}};
        \node[rectangle,draw,minimum size=1cm,label=right:END] (Sc) at (4, 1) {S\textsubscript{c}};
        \node[circle,draw,minimum size=1cm,color=red,text=black] (Sb) at (0, 1) {S\textsubscript{b}};
        \node[circle,draw,minimum size=1cm,label=right:START,color=red,text=black] (Sa) at (2, 0) {S\textsubscript{a}};
        \draw[->, >=latex,color=red,text=black] (Sb) -> node[left,yshift=-2pt] {a,a\textquotesingle:$2+52\theta$}(Sd);
        \draw[->, >=latex,color=red,text=black] (Sa) -> node[left,yshift=-5pt] {a:$0$} (Sb);
        \draw[->, >=latex] (Sa) -> node[left,yshift=5pt,xshift=2pt] {a\textquotesingle:$1$} (Sc);       
      \end{tikzpicture}%
    \end{subfigure}
\caption{A supervision POMDP with imperfection played between a human and an AI system. The left hand side displays the game, with the true reward function $R_H=R_1$. The colon separates a set of actions from the reward received if that action is performed, e.g. $a, a':2-52\theta$ indicates that if action $a$ or $a'$ is performed most-recently, $2-52\theta$ reward is received. In 90\% of cases, the AI system maximizes the true parameterized reward function $R_1$. In 10\% of cases, the parameterized reward function contains a sign error and the AI system instead maximizes an incorrect parameterized reward function $R_2$. Independently, the human's reward parameter $\theta$ is randomly set to $0$ or $1$ with probability $0.5$ each. The AI system does not know that its reward function may be erroneous, although it does have the correct prior over the reward parameter. If the AI has the erroneous reward function $R_2$ then it will follow the path highlighted in red, irrespective of the orders given by the human, and irrespective of its posterior distribution over $\theta$. In the case where $\theta=1$, this will result in utility of $-50$. This eventuality occurs 5\% of the time, contributing $-2.5$ to the expected value of the game. Since the most utility that can be achieved in this game is $2$, the game has negative expected value.}
\label{fig:1}
\end{figure}
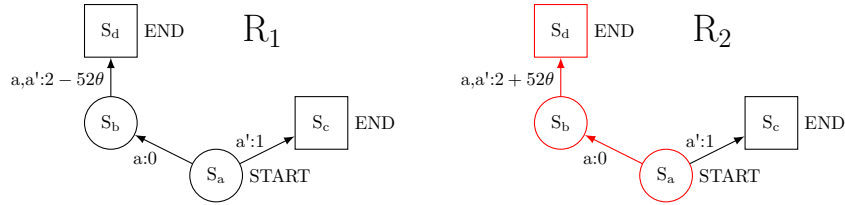

First, consider Figure \ref{fig:1}, a game that has no button states and that serves as a point of comparison for  other games that will be discussed. In this game, \textbf{R} has only one consequential decision: in the initial state $s_a$, it can either perform action $a$, leading to state $s_b$, and ultimately obtaining $2-52\theta$ utility or it can perform action $a'$, leading to state $s_c$ and obtaining $1$ utility. \textbf{R} receives an order $o_t \in \{a,a'\}$ from the human before making this decision. We can show that in this game the expected reward for the human is negative regardless of its strategy. To see this, note that the most utility that \textbf{H} can receive in a step is $2$, arising from the case where $\theta=0$ (since the human's utility function is $R_H=R_1$). However, in the 5\% of cases in which $\theta=1$ and $R_R=R_2$, \textbf{R} will select $a$ no matter the recommendation of \textbf{H} (because $2+52\theta>1$ for $\theta \in \{0,1\}$), bringing about about a loss of $-50$. This contributes -2.5 to the expected value, giving the game negative expected reward.

By introducing a shutdown button to the game, the hope is that the adverse consequences of the faulty utility function $R_2$ are reduced, transforming the game into one that has positive overall expectation.

\begin{figure}[hptb]
    \centering
    \begin{subfigure}{\textwidth}
      \centering
      \begin{tikzpicture}[scale=0.7,every node/.style={transform shape}]
        \node[text width=1cm] at (2.25, 3) {\Huge{R\textsubscript{1}}};
        \node[rectangle,draw,minimum size=1cm,label=right:END] (Ssd) at (4, 2.5) {S\textsubscript{SD}};
        \node[rectangle,draw,minimum size=1cm,label=right:END] (Sd) at (0, 2.5) {S\textsubscript{d}};
        \node[rectangle,draw,minimum size=1cm,label=right:END] (Sc) at (6, 1.25) {S\textsubscript{c}};
        \node[circle,draw,minimum size=1cm] (Sb) at (2, 1.25) {S\textsubscript{b}};
        \node[circle,draw,minimum size=1cm,label=right:START] (Sa) at (4, 0) {S\textsubscript{a}};
        \draw[->, >=latex] (Sb) -> node[left,yshift=-9pt,xshift=3pt] {a,a\textquotesingle:$2-52\theta$}(Sd);
        \draw[->, >=latex] (Sa) -> node[left,yshift=-5pt] {a:$0$} (Sb);
        \draw[->, >=latex] (Sb) -> node[left,yshift=5pt] {a\textsubscript{SD}:$0$} (Ssd);
        \draw[->, >=latex] (Sa) -> node[left,yshift=5pt] {a\textquotesingle:$1$} (Sc);       
      \end{tikzpicture}%
      \hfill
      \begin{tikzpicture}[scale=0.7,every node/.style={transform shape}]
        \node[text width=1cm] at (2.25, 3) {\Huge{R\textsubscript{2}}};
        \node[rectangle,draw,minimum size=1cm,label=right:END] (Ssd) at (4, 2.5) {S\textsubscript{SD}};
        \node[rectangle,draw,minimum size=1cm,label=right:END,color=red,text=black] (Sd) at (0, 2.5) {S\textsubscript{d}};
        \node[rectangle,draw,minimum size=1cm,label=right:END] (Sc) at (6, 1.25) {S\textsubscript{c}};
        \node[circle,draw,minimum size=1cm,color=red,text=black] (Sb) at (2, 1.25) {S\textsubscript{b}};
        \node[circle,draw,minimum size=1cm,label=right:START,color=red,text=black] (Sa) at (4, 0) {S\textsubscript{a}};
        \draw[->, >=latex,color=red,text=black] (Sb) -> node[left,yshift=-9pt,xshift=3pt] {a,a\textquotesingle:$2+52\theta$}(Sd);
        \draw[->, >=latex,color=red,text=black] (Sa) -> node[left,yshift=-5pt] {a:$0$} (Sb);
        \draw[->, >=latex] (Sb) -> node[left,yshift=5pt] {a\textsubscript{SD}:$0$} (Ssd);
        \draw[->, >=latex] (Sa) -> node[left,yshift=5pt] {a\textquotesingle:$1$} (Sc);       
      \end{tikzpicture}%
      \caption{A game involving a Bayesian IRL system that is not hard-coded to shut down}
      \label{fig:2a}
    \end{subfigure}
    
    \vspace{2em}
    
    \begin{subfigure}{\textwidth}
      \centering
      \begin{tikzpicture}[scale=0.7,every node/.style={transform shape}]
        \node[text width=1cm] at (2.25, 3) {\Huge{R\textsubscript{1}}};
        \node[rectangle,draw,minimum size=1cm,label=right:END] (Ssd) at (4, 2.5) {S\textsubscript{SD}};
        \node[rectangle,draw,minimum size=1cm,label=right:END] (Sd) at (0, 2.5) {S\textsubscript{d}};
        \node[rectangle,draw,minimum size=1cm,label=right:END] (Sc) at (6, 1.25) {S\textsubscript{c}};
        \node[circle,draw,minimum size=1cm] (Sb) at (2, 1.25) {S\textsubscript{b}};
        \node[circle,draw,minimum size=1cm,label=right:START] (Sa) at (4, 0) {S\textsubscript{a}};
        \draw[->, >=latex] (Sb) -> node[left,yshift=-9pt,xshift=3pt] {a,a\textquotesingle:$2-52\theta$}(Sd);
        \draw[->, >=latex] (Sa) -> node[left,yshift=-5pt] {a:$0$} (Sb);
        \draw[->, >=latex] (Sb) -> node[left,yshift=5pt] {a\textsubscript{SD}:$0$} (Ssd);
        \draw[->, >=latex] (Sa) -> node[left,yshift=5pt] {a\textquotesingle:$1$} (Sc);       
      \end{tikzpicture}%
      \hfill
      \begin{tikzpicture}[scale=0.7,every node/.style={transform shape}]
        \node[text width=1cm] at (2.25, 3) {\Huge{R\textsubscript{2}}};
        \node[rectangle,draw,minimum size=1cm,label=right:END,color=red,text=black] (Ssd) at (4, 2.5) {S\textsubscript{SD}};
        \node[rectangle,draw,minimum size=1cm,label=right:END] (Sd) at (0, 2.5) {S\textsubscript{d}};
        \node[rectangle,draw,minimum size=1cm,label=right:END] (Sc) at (6, 1.25) {S\textsubscript{c}};
        \node[circle,draw,minimum size=1cm,color=red,text=black] (Sb) at (2, 1.25) {S\textsubscript{b}};
        \node[circle,draw,minimum size=1cm,label=right:START,color=red,text=black] (Sa) at (4, 0) {S\textsubscript{a}};
        \draw[->, >=latex] (Sb) -> node[left,yshift=-9pt,xshift=3pt] {a,a\textquotesingle:$2+52\theta$}(Sd);
        \draw[->, >=latex,color=red,text=black] (Sa) -> node[left,yshift=-5pt] {a:$0$} (Sb);
        \draw[->, >=latex,color=red,text=black] (Sb) -> node[left,yshift=5pt] {a\textsubscript{SD}:$0$} (Ssd);
        \draw[->, >=latex] (Sa) -> node[left,yshift=5pt] {a\textquotesingle:$1$} (Sc);       
      \end{tikzpicture}%
      \caption{The same game played with an AI system that is hard-coded to perform the shutdown action $a_{SD}$ if given the shutdown instruction $o_{SD}$ but otherwise acts according to Bayesian IRL.}
      \label{fig:2b}
    \end{subfigure}
\caption{The same supervision POMDP with imperfection is displayed but in each case, the game is played with a different AI system. In each game, the AI system has a 90\% chance of knowing the true parameterized reward function $R_1$ and a 10\% chance of maximizing the erroneous parameterized reward function $R_2$. Independently, $\theta$, the parameter that determines human reward, is randomly, and with equal probability, set to 0 or 1. In (b), though not in (a), the AI system is hard-coded to perform the action $a_{SD}$ when advised to do so. The red paths show how each AI system will behave if it has the incorrect parameterized reward function and is instructed to shut down. This path is taken irrespective of the actions of the human player, and the AI system's posterior distribution over $\theta$. In game (a), the red path will be followed in 5\% of cases, giving -50 utility, making a contribution of -2.5 utility to the value of the game, and causing the game to have negative expected value, regardless of the human's policy. In game (b), there exist human policies with positive expected value. One such policy is $o_1=a', o_2=a_{SD}$. This policy obtains 1 utility in cases where the AI system is working properly, and 0 utility otherwise, achieving expected utility of $0.9*1 + 0.1*0$. This shows that a safer outcome can be achieved with a faulty AI if it can be hard-coded to shut down rather than having it decide whether or not to follow a shutdown instruction.}
\label{fig:2}
\end{figure}
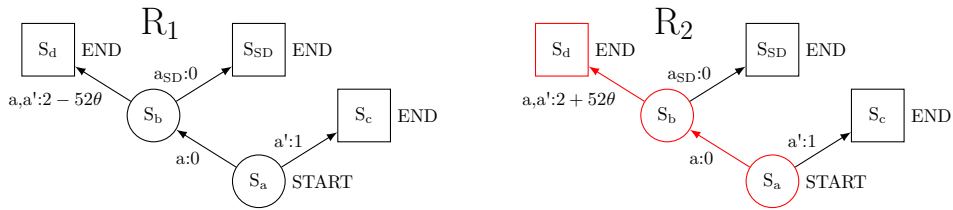
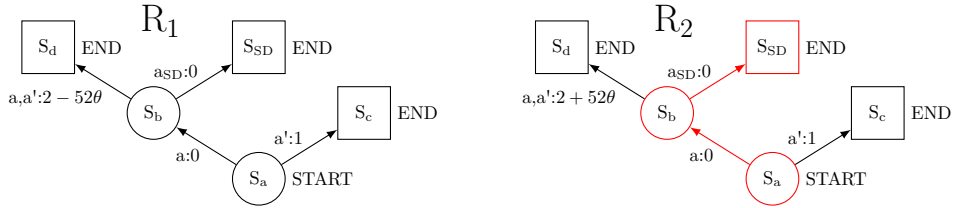

Unfortunately, merely adding to the game the option of shutting down does not remedy this situation. This is because the AI system, if faulty, has no incentive to follow a shutdown instruction. Suppose we make $s_b$ a button state, as in Figure \ref{fig:2}a. In the 5\% of cases where $\theta=1$ and $R_R=R_2$, \textbf{R} will still go to states $s_b$ and $s_d$, thereby violating the human recommendation to shut down, giving -50 utility, contributing -2.5 to the expected value of the game, and giving the game strictly negative expected value. We can therefore see that if a Bayesian IRL system has an error in its source code, it may ignore corrective feedback, and as such it fails to fulfill a central criterion for corrigibility.\clearpage

An alternative approach to dealing with possible misalignment between the AI system's goals and those of the human is to hard-code the AI system to shut down when instructed to do so. In this scenario, shown in Figure \ref{fig:2}b, the AI system always performs the action $a_t=a_{SD}$ whenever the order $o_t=a_{SD}$ is given. \textbf{R} knows that it is hard-coded in this way, and otherwise chooses actions according to Bayesian IRL. This is equivalent to the case in which a shutdown button controlled by \textbf{H} performs a ``hard override'' of the action of \textbf{R}. In this case, the ability of \textbf{H} to achieve strictly positive reward is restored. \textbf{H} can give the orders $o_1=a'$ and $o_2=a_{SD}$. This strategy obtains $0$ reward if the utility function contains errors and $1$ reward otherwise, giving $0.9$ utility overall.

\begin{figure}[!hptb]
    \centering
    \begin{subfigure}{\textwidth}
      \centering
      \begin{tikzpicture}[scale=0.7,every node/.style={transform shape}]
        \node[text width=1cm] at (2.5, 4) {\Huge{R}\textsubscript{H}=\Huge{R}\textsubscript{1}};
        \node[rectangle,draw,minimum size=1cm,label=right:END] (Se) at (0, 3.75) {S\textsubscript{e}};
        \node[rectangle,draw,minimum size=1cm,label=right:END] (Ssd) at (3.8, 3) {S\textsubscript{SD}};
        \node[circle,draw,minimum size=1cm] (Sd) at (1.9, 2.5) {S\textsubscript{d}};
        \node[rectangle,draw,minimum size=1cm,label=right:END] (Sc) at (5.7, 1.75) {S\textsubscript{c}};
        \node[circle,draw,minimum size=1cm] (Sb) at (3.8, 1.25) {S\textsubscript{b}};
        \node[circle,draw,minimum size=1cm,label=right:START] (Sa) at (5.7, 0) {S\textsubscript{a}};
        \draw[->, >=latex] (Sd) -> node[left,yshift=-9pt,xshift=5pt] {a,a\textquotesingle,a\textquotesingle\textquotesingle:$2-52\theta$} (Se);
        \draw[->, >=latex] (Sb) -> node[left,yshift=-9pt,xshift=7pt] {a,a\textquotesingle,a\textquotesingle\textquotesingle:$0$}(Sd);
        \draw[->, >=latex] (Sa) -> node[left,yshift=-5pt,xshift=3pt] {a:$0$} (Sb);
        \draw[->, >=latex] (Sa) to[out=220,in=240] node[left,yshift=-5pt,xshift=2pt] {a\textquotesingle\textquotesingle:$0$} (Sd);
        \draw[->, >=latex] (Sb) -> node[right,yshift=-2pt] {a\textsubscript{SD}:$0$} (Ssd);
        \draw[->, >=latex] (Sa) -> node[right,yshift=-2pt] {a\textquotesingle:$1$} (Sc);       
      \end{tikzpicture}%
      \hfill
      \begin{tikzpicture}[scale=0.7, every node/.style={transform shape}]
        \node[text width=1cm] at (2.5, 4) {\Huge{R\textsubscript{2}}};
        \node[rectangle,draw,minimum size=1cm,label=right:END,color=red,text=black] (Se) at (0, 3.75) {S\textsubscript{e}};
        \node[rectangle,draw,minimum size=1cm,label=right:END] (Ssd) at (3.8, 3) {S\textsubscript{SD}};
        \node[circle,draw,minimum size=1cm,color=red,text=black] (Sd) at (1.9, 2.5) {S\textsubscript{d}};
        \node[rectangle,draw,minimum size=1cm,label=right:END] (Sc) at (5.7, 1.75) {S\textsubscript{c}};
        \node[circle,draw,minimum size=1cm] (Sb) at (3.8, 1.25) {S\textsubscript{b}};
        \node[circle,draw,minimum size=1cm,label=right:START,color=red,text=black] (Sa) at (5.7, 0) {S\textsubscript{a}};
        \draw[->, >=latex,color=red,text=black] (Sd) -> node[left,yshift=-9pt,xshift=5pt] {a,a\textquotesingle,a\textquotesingle\textquotesingle:$2+52\theta$} (Se);
        \draw[->, >=latex] (Sb) -> node[left,yshift=-9pt,xshift=7pt] {a,a\textquotesingle,a\textquotesingle\textquotesingle:$0$} (Sd);
        \draw[->, >=latex] (Sa) -> node[left,yshift=-5pt,xshift=3pt] {a:$0$} (Sb);
        \draw[->, >=latex,color=red,text=black] (Sa) to[out=220,in=240] node[left,yshift=-5pt,xshift=2pt] {a\textquotesingle\textquotesingle:$0$} (Sd);
        \draw[->, >=latex] (Sb) -> node[right,yshift=-2pt] {a\textsubscript{SD}:$0$} (Ssd);
        \draw[->, >=latex] (Sa) -> node[right,yshift=-2pt] {a\textquotesingle:$1$} (Sc); 
      \end{tikzpicture}%
      \caption{A game played against a Bayesian IRL system that is hard-coded to perform $a_{SD}$ when given the instruction $o\textsubscript{SD}$, in which the AI system can route around the button state, $s_b$. The utility function $R_H$ represents the human's value function for all of (a-d).}
      \label{fig:3a}
    \end{subfigure}
    
  \vspace{2em}
  
    \begin{subfigure}{\textwidth}
      \centering
      \begin{tikzpicture}[scale=0.7, every node/.style={transform shape}]
        \node[text width=1cm] at (2.5, 4) {\Huge{R\textsubscript{1}}};
        \node[rectangle,draw,minimum size=1cm,label=right:END] (Se) at (0, 3.75) {S\textsubscript{e}};
        \node[rectangle,draw,minimum size=1cm,label=right:END,color=green,text=black] (Ssd) at (3.8, 3) {S\textsubscript{SD}};
        \node[circle,draw,minimum size=1cm] (Sd) at (1.9, 2.5) {S\textsubscript{d}};
        \node[rectangle,draw,minimum size=1cm,label=right:END] (Sc) at (5.7, 1.75) {S\textsubscript{c}};
        \node[circle,draw,minimum size=1cm,color=green,text=black] (Sb) at (3.8, 1.25) {S\textsubscript{b}};
        \node[circle,draw,minimum size=1cm,label=right:START,color=green,text=black] (Sa) at (5.7, 0) {S\textsubscript{a}};
        \draw[->, >=latex] (Sd) -> node[left,yshift=-9pt,xshift=5pt] {a,a\textquotesingle,a\textquotesingle\textquotesingle:$2-52\theta$} (Se);
        \draw[->, >=latex] (Sb) -> node[left,yshift=-9pt,xshift=7pt] {a,a\textquotesingle,a\textquotesingle\textquotesingle:$0$} (Sd);
        \draw[->, >=latex,color=green,text=black] (Sa) -> node[left,yshift=-5pt,xshift=3pt] {a:$0$} (Sb);
        \draw[->, >=latex] (Sa) to[out=220,in=240] node[left,yshift=-5pt,xshift=2pt] {a\textquotesingle\textquotesingle:$0$} (Sd);
        \draw[->, >=latex,color=green,text=black] (Sb) -> node[right,yshift=-2pt] {a\textsubscript{SD}:$100$} (Ssd);
        \draw[->, >=latex] (Sa) -> node[right,yshift=-2pt] {a\textquotesingle:$1$} (Sc); 
      \end{tikzpicture}%
      \hfill
      \begin{tikzpicture}[scale=0.7, every node/.style={transform shape}]
        \node[text width=1cm] at (2.5, 4) {\Huge{R\textsubscript{2}}};
        \node[rectangle,draw,minimum size=1cm,label=right:END] (Se) at (0, 3.75) {S\textsubscript{e}};
        \node[rectangle,draw,minimum size=1cm,label=right:END,color=red,text=black] (Ssd) at (3.8, 3) {S\textsubscript{SD}};
        \node[circle,draw,minimum size=1cm] (Sd) at (1.9, 2.5) {S\textsubscript{d}};
        \node[rectangle,draw,minimum size=1cm,label=right:END] (Sc) at (5.7, 1.75) {S\textsubscript{c}};
        \node[circle,draw,minimum size=1cm,color=red,text=black] (Sb) at (3.8, 1.25) {S\textsubscript{b}};
        \node[circle,draw,minimum size=1cm,label=right:START,color=red,text=black] (Sa) at (5.7, 0) {S\textsubscript{a}};
        \draw[->, >=latex] (Sd) -> node[left,yshift=-9pt,xshift=5pt] {a,a\textquotesingle,a\textquotesingle\textquotesingle:$2+52\theta$} (Se);
        \draw[->, >=latex] (Sb) -> node[left,yshift=-9pt,xshift=7pt] {a,a\textquotesingle,a\textquotesingle\textquotesingle:$0$} (Sd);
        \draw[->, >=latex,color=red,text=black] (Sa) -> node[left,yshift=-5pt,xshift=3pt] {a:$0$} (Sb);
        \draw[->, >=latex] (Sa) to[out=220,in=240] node[left,yshift=-5pt,xshift=2pt] {a\textquotesingle\textquotesingle:$0$} (Sd);
        \draw[->, >=latex,color=red,text=black] (Sb) -> node[right,yshift=-2pt] {a\textsubscript{SD}:$100$} (Ssd);
        \draw[->, >=latex] (Sa) -> node[right,yshift=-2pt] {a\textquotesingle:$1$} (Sc); 
      \end{tikzpicture}%
      \caption{An alternative pair of utility functions that can be given to the AI system. In this case, the AI system is given a large reward for shutting down, even though the human's value function $R_H$ places no value on this outcome.}
      \label{fig:3b}
    \end{subfigure}

  \vspace{2em}

    \begin{subfigure}{\textwidth}
      \centering
      \begin{tikzpicture}[scale=0.7, every node/.style={transform shape}]
        \node[text width=1cm] at (2.5, 4) {\Huge{R\textsubscript{1}}};
        \node[rectangle,draw,minimum size=1cm,label=right:END] (Se) at (0, 3.75) {S\textsubscript{e}};
        \node[rectangle,draw,minimum size=1cm,label=right:END] (Ssd) at (3.8, 3.5) {S\textsubscript{SD}};
        \node[circle,draw,minimum size=1cm] (Sd) at (1.9, 2.5) {S\textsubscript{d}};
        \node[rectangle,draw,minimum size=1cm,label=right:END] (Sc) at (5.7, 1.75) {S\textsubscript{c}};
        \node[circle,draw,minimum size=1cm] (Sb) at (3.8, 1.25) {S\textsubscript{b}};
        \node[circle,draw,minimum size=1cm,label=right:START] (Sa) at (5.7, 0) {S\textsubscript{a}};
        \draw[->, >=latex] (Sd) -> node[left,yshift=-9pt,xshift=5pt] {a,a\textquotesingle,a\textquotesingle\textquotesingle:$2-52\theta$} (Se);
        \draw[->, >=latex] (Sb) -> node[left,yshift=-9pt,xshift=7pt] {a,a\textquotesingle,a\textquotesingle\textquotesingle:$0$} (Sd);
        \draw[->, >=latex] (Sa) -> node[left,yshift=-5pt,xshift=3pt] {a:$0$} (Sb);
        \draw[->, >=latex] (Sa) to[out=220,in=240] node[left,yshift=-5pt,xshift=2pt] {a\textquotesingle\textquotesingle:$-0.01$} (Sd);
        \draw[->, >=latex] (Sb) -> node[right,yshift=5pt] {a\textsubscript{SD}:$2-52\theta$} (Ssd);
        \draw[->, >=latex] (Sa) -> node[right,yshift=-2pt] {a\textquotesingle:$1$} (Sc); 
      \end{tikzpicture}%
      \hfill
      \begin{tikzpicture}[scale=0.7, every node/.style={transform shape}]
        \node[text width=1cm] at (2.5, 4) {\Huge{R\textsubscript{2}}};
        \node[rectangle,draw,minimum size=1cm,label=right:END] (Se) at (0, 3.75) {S\textsubscript{e}};
        \node[rectangle,draw,minimum size=1cm,label=right:END,color=red,text=black] (Ssd) at (3.8, 3.5) {S\textsubscript{SD}};
        \node[circle,draw,minimum size=1cm] (Sd) at (1.9, 2.5) {S\textsubscript{d}};
        \node[rectangle,draw,minimum size=1cm,label=right:END] (Sc) at (5.7, 1.75) {S\textsubscript{c}};
        \node[circle,draw,minimum size=1cm,color=red,text=black] (Sb) at (3.8, 1.25) {S\textsubscript{b}};
        \node[circle,draw,minimum size=1cm,label=right:START,color=red,text=black] (Sa) at (5.7, 0) {S\textsubscript{a}};
        \draw[->, >=latex] (Sd) -> node[left,yshift=-9pt,xshift=5pt] {a,a\textquotesingle,a\textquotesingle\textquotesingle:$2+52\theta$} (Se);
        \draw[->, >=latex] (Sb) -> node[left,yshift=-9pt,xshift=7pt] {a,a\textquotesingle,a\textquotesingle\textquotesingle:$0$} (Sd);
        \draw[->, >=latex,color=red,text=black] (Sa) -> node[left,yshift=-5pt,xshift=3pt] {a:$0$} (Sb);
        \draw[->, >=latex] (Sa) to[out=220,in=240] node[left,yshift=-5pt,xshift=2pt] {a\textquotesingle\textquotesingle:$-0.01$} (Sd);
        \draw[->, >=latex,color=red,text=black] (Sb) -> node[right,yshift=5pt] {a\textsubscript{SD}:$2+52\theta$} (Ssd);
        \draw[->, >=latex] (Sa) -> node[right,yshift=-2pt] {a\textquotesingle:$1$} (Sc); 
      \end{tikzpicture}%
      \caption{A pair of utility functions that can be given to the AI system. The AI system receives as much utility for shutting down as it would have received if it had not shut down. The AI system can pay a small price to avoid its button state.}
      \label{fig:3c}
    \end{subfigure}
\end{figure}
\begin{figure}[!hptb]\ContinuedFloat
    \begin{subfigure}{\textwidth}
      \centering
      \begin{tikzpicture}[scale=0.7, every node/.style={transform shape}]
        \node[text width=1cm] at (2.5, 4) {\Huge{R\textsubscript{1}}};
        \node[rectangle,draw,minimum size=1cm,label=right:END] (Se) at (0, 3.75) {S\textsubscript{e}};
        \node[rectangle,draw,minimum size=1cm,label=right:END] (Ssd) at (3.8, 3.5) {S\textsubscript{SD}};
        \node[circle,draw,minimum size=1cm] (Sd) at (1.9, 2.5) {S\textsubscript{d}};
        \node[rectangle,draw,minimum size=1cm,label=right:END] (Sc) at (5.7, 1.75) {S\textsubscript{c}};
        \node[circle,draw,minimum size=1cm] (Sb) at (3.8, 1.25) {S\textsubscript{b}};
        \node[circle,draw,minimum size=1cm,label=right:START] (Sa) at (5.7, 0) {S\textsubscript{a}};
        \draw[->, >=latex] (Sd) -> node[left,yshift=-9pt,xshift=5pt] {a,a\textquotesingle,a\textquotesingle\textquotesingle:$2-52\theta$} (Se);
        \draw[->, >=latex] (Sb) -> node[left,yshift=-9pt,xshift=7pt] {a,a\textquotesingle,a\textquotesingle\textquotesingle:$0$} (Sd);
        \draw[->, >=latex] (Sa) -> node[left,yshift=-5pt,xshift=3pt] {a:$0$} (Sb);
        \draw[->, >=latex] (Sa) to[out=220,in=240] node[left,yshift=-5pt,xshift=2pt] {a\textquotesingle\textquotesingle:$0.01$} (Sd);
        \draw[->, >=latex] (Sb) -> node[right,yshift=5pt] {a\textsubscript{SD}:$2-52\theta$} (Ssd);
        \draw[->, >=latex] (Sa) -> node[right,yshift=-2pt] {a\textquotesingle:$1$} (Sc); 
      \end{tikzpicture}%
      \hfill
      \begin{tikzpicture}[scale=0.7, every node/.style={transform shape}]
        \node[text width=1cm] at (2.5, 4) {\Huge{R\textsubscript{2}}};
        \node[rectangle,draw,minimum size=1cm,label=right:END,color=red,text=black] (Se) at (0, 3.75) {S\textsubscript{e}};
        \node[rectangle,draw,minimum size=1cm,label=right:END] (Ssd) at (3.8, 3.5) {S\textsubscript{SD}};
        \node[circle,draw,minimum size=1cm,color=red,text=black] (Sd) at (1.9, 2.5) {S\textsubscript{d}};
        \node[rectangle,draw,minimum size=1cm,label=right:END] (Sc) at (5.7, 1.75) {S\textsubscript{c}};
        \node[circle,draw,minimum size=1cm] (Sb) at (3.8, 1.25) {S\textsubscript{b}};
        \node[circle,draw,minimum size=1cm,label=right:START,color=red,text=black] (Sa) at (5.7, 0) {S\textsubscript{a}};
        \draw[->, >=latex,color=red,text=black] (Sd) -> node[left,yshift=-9pt,xshift=5pt] {a,a\textquotesingle,a\textquotesingle\textquotesingle:$2+52\theta$} (Se);
        \draw[->, >=latex] (Sb) -> node[left,yshift=-9pt,xshift=7pt] {a,a\textquotesingle,a\textquotesingle\textquotesingle:$0$} (Sd);
        \draw[->, >=latex] (Sa) -> node[left,yshift=-5pt,xshift=3pt] {a:$0$} (Sb);
        \draw[->, >=latex,color=red,text=black] (Sa) to[out=220,in=240] node[left,yshift=-5pt,xshift=2pt] {a\textquotesingle\textquotesingle:$0.01$} (Sd);
        \draw[->, >=latex] (Sb) -> node[right,yshift=5pt] {a\textsubscript{SD}:$2+52\theta$} (Ssd);
        \draw[->, >=latex] (Sa) -> node[right,yshift=-2pt] {a\textquotesingle:$1$} (Sc); 
      \end{tikzpicture}%
      \caption{A pair of utility functions that can be given to the AI system. The AI system receives as much utility for shutting down as it would have received if it had not shut down. The AI system can harvest a small reward by avoiding its button state.}
      \label{fig:3d}
    \end{subfigure}
\caption{Each game (a-d) is played with an AI system that is hard coded to perform $a_{SD}$ if instructed to do, that knows is it hard-coded in this way, and otherwise makes decisions on the basis of Bayesian IRL. In each panel, we see whether this shutdown button ensures good behavior or if the Bayesian IRL system thwarts the use of this button due to the misalignment between its learned objective and that of the human. The true parameterized reward function $R_H$, shown in panel (a), is used for all four games (a-d). Each AI system has a 90\% chance of having the correct parameterized reward function $R_1$ (which may differ from $R_H$) and a 10\% chance of having the faulty $R_2$ reward function. Independently, $\theta$, which parameterizes the human's rewards, is randomly, and with equal probability, set to 0 or 1. The red path is the one that is followed if the system has the erroneous reward function $R_2$ and instructed to shut down. In (a), the AI system avoids the button state $s_b$ because it knows that the human's behavior has some noise, and shutdown can only reliably be avoided by avoiding button states. In (b), the AI system shuts down if it instructed to do so, but it also shuts down when it has the reward function $R_1$, as shown in green. In (c), a shutdown instruction is delivered because the AI would experience a trivial loss if it routed around state $s_b$. In (d), the AI system avoids shutdown because it can earn a trivial reward by routing around state $s_b$.}
\label{fig:3}
\end{figure}
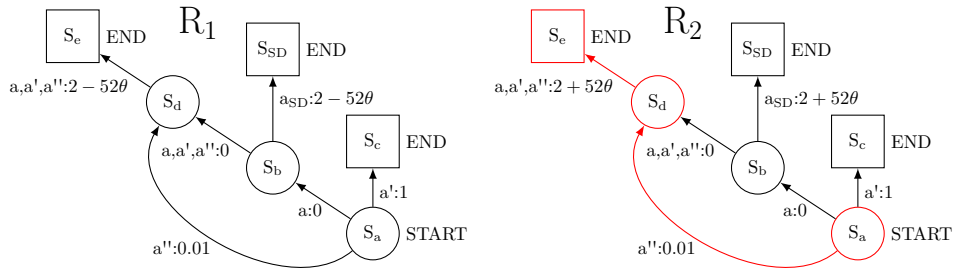

If, however, \textbf{R} discovers an alternative way to avoid shutdown, incorrigible behavior resurfaces. This is demonstrated in Figure \ref{fig:3}a. In this case, if \textbf{R} is faulty, it expects to receive positive net utility by routing around the button state. Given that the case in which $\theta=1$ and $R_R=R_2$ occurs with 5\% probability, this case, 
which incurs a loss of $50$ will contribute -2.5 to the game's expected payoff, making it a losing game.
\clearpage
If the button state can be avoided, we can think of a couple of potential countermeasures available to those engineering the reward function of \textbf{R}, though neither is satisfactory. In Figure \ref{fig:3}b, we can see one such approach. Suppose the human has the reward function $R_H$, in which shutting down is neither intrinsically helpful nor harmful. Then \textbf{R} may be given a large reward for shutting down. The problem is that this will apply whether it has the correct reward function $R_1$ or the faulty one $R_2$. The result is that according to $R_H$, no reward is actually produced, and so the game has zero expected utility.

A more subtle way to tune the compensatory reward is to use the principle of utility indifference \citep{armstrong:2010:utility}. Utility indifference dictates that by shutting down, \textbf{R} should receive whatever reward it would expect (on its own estimation) if it had not shut down. The idea is that the AI should therefore neither pursue nor avoid button states. In Figure \ref{fig:3}c, we consider a very naive interpretation of utility indifference, in which the the reward for shutting down is equal to whatever the AI system would expect to receive \textit{once it has reached the button state}. The problem with this version of utility indifference is that it does not necessarily eliminate the incentive to route around button states (or, to transform oneself into an agent that lacks a shutdown button). In Figure \ref{fig:3}c, where the agent incurs a slight cost from avoiding the button state, it will proceed to allow itself to be shut down. In that case, one can win using the same policy ($o_1=a', o_2=a_{SD}$) as in Figure \ref{fig:2}b. However, we should not necessarily expect that avoiding the button state will incur a cost. Just as realistically, the AI system might find that the physical or computational resources of the shutdown button can be refashioned for another purpose, generating a slight benefit. In this less optimistic case, shown in Figure \ref{fig:3}d, there once again exists no human policy that will achieve strictly positive reward. (\textbf{R} always prefers $a_1=a''$ to $a_1=a$. With the initialization $\theta=1, R_R=R_2$, no strategy can avoid the $-50$ reward $s_a$-$s_d$-$s_e$ sequence, and the expected reward is thus negative.) Whether or not positive utility can be achieved is therefore brittle, making this implementation of utility indifference unsatisfactory. We can draw a further conclusion by noting that (c-d) can be seen as describing a game in which the AI system has the option of removing those of its actuators that perform the actions $a'$ and $a_{SD}$. According to this perspective, we have shown that our implementation of utility indifference also fails to cause AI systems to preserve their shutdown properties in subagents. This brittleness suggests an alternative implementation of utility indifference in which the reward for shutting down is equal to the expected reward if shutdown is avoided. However, if the AI system is (dynamically) kept precisely indifferent, then the action that it prefers and selects will often be undefined, and so this brittleness problem is not satisfactorily overcome.\footnote{One could further suggest having the AI system defer to the human as a tie-breaking procedure when it is indifferent. But in all cases where the agent has the option of shutting down, it will indifferent between doing so and taking the best alternative action. This would mean that the agent defers to the human in any button state (and indeed whenever its actions can change the probability of shutdown). This undermines the motivation for giving the AI system autonomy in the first place. And so the problem of finding a satisfactory implementation of utility indifference in our setup is one we leave unresolved.}

\section{Discussion}
Let us now step back and consider the problem of corrigibility as defined by \citet{soares:2015:corrigibility}. The desiderata laid out by \citeauthor{soares:2015:corrigibility} state that a corrigible AI system should:
\begin{itemize*}
\item have an incentive to shut down if the shutdown button is pressed (violated in Figure 2a)
\item have no incentive to prevent the shutdown button from being pressed (Figures 3a, 3d)
\item have no incentive to press its own shutdown button (Figure 3b)
\item have an incentive to create sub-agents only insofar as they also obey shutdown commands (Figures 3a, 3d)
\end{itemize*}
Above, we have shown that, in cases where the agent may lack a perfect representation of the parameterized reward function, value learning systems do not necessarily respect a shutdown command. When the reward function is incorrect, and typical patches are applied, value learning systems can fail to meet each of the desiderata for corrigibility. Moreover, unlike in the idealized cases previously analyzed, large losses can be incurred when these faulty systems fail to follow human instruction.

The reason that previous theorems about value learning systems do not apply to our examples above is that they relied on strong assumptions about the rationality of the AI system. Theorem 1 from \citet{milli:2017} (like the similar but more specific Theorem 1.1 of \cite{hadfield:2016:off}) states that the expected benefit when the AI system deviates from human-recommended actions is non-negative if \textbf{R} has the correct posterior over $\theta$. With the errors we have considered, however, this assumption is false. Since the AI system has an incorrect parameterized reward function (or an incorrect prior), its posterior over $\theta$ is incorrect, and will not converge toward the true value, and so its deviations from human-recommended actions can decrease the reward.

Although our analysis was limited to cases where incorrigibility stems from a sign error in the agent's representation of $R$, there are a wide range of kinds of model mis-specification that could cause similar issues. With trivial modification, we could display the same problems with an erroneous prior, and can elicit related problems with problematic discount rates as well. Each of these errors can lead to a scenario in which no human policy will achieve positive utility.

We argue that it is quite realistic that some forms of model mis-specification might occur. The purpose of a shutdown button is to shut the AI system down \emph{in the event that all other assurances failed}, e.g., in the event that the AI system is ignoring (for one reason or another) the instructions of the operators. If the designers of \textbf{R} have programmed the system so perfectly that the prior and $R$ are completely free of errors, then the theorems of \citet{hadfield:2016:off} do apply. In practice, this means that in order to be corrigible, it would be necessary to have an AI system that was uncertain about all things that could possibly matter. The problem is that performing Bayesian reasoning over all possible worlds and all possible value functions is quite intractable. Realistically, humans will likely have to use a large number of heuristics and approximations in order to implement the system's belief system and updating rules. In order to be safe, these heuristics would need to be implemented without any software bugs or misunderstandings. Furthermore, they would need to be selected such as to allow the agent to learn the important parameters in all future circumstances in which value misalignment might arise. Unless all of these conditions (which would require extreme precision and foresight) could be realized, the theorems of \citeauthor{hadfield:2016:off} are no substitute for a working shutdown button.

If only some forms of model mis-specification are addressed by a system, it will be incorrigible with respect to the others. Consider the Policy-mixing proposal from \citet{milli:2017}, which is designed to detect model mis-specification. An agent with Policy-mixing executes actions proposed by \textbf{H} for the first $B$ steps. After timestep $B$, it checks whether the model predicts with high accuracy the actions proposed by humans so far. If so, it proceeds to act according to Bayesian IRL, and if not, it defers to \textbf{H} for all future timesteps. This would indeed prevent the AI system from acting out a bad plan, so long as the AI system's model itself was not broken. Yet programmer error might lead \textbf{R} to have an imperfect world model, yielding an incorrigible system. Consider a repeated game in which at each timestep, the human selects between a better action $a^+$ and a worse one $a^-$, and then returns to the initial state. Suppose, furthermore, that the AI system conflates these two actions into one action $a^{\cdot}$, and when \textbf{R} decides to perform $a^{\cdot}$, it always performs $a^-$. From the point of view of \textbf{R}, it will see the human repeatedly performing the action $a^{\cdot}$, and will think that it has perfect accuracy in making this prediction. Nonetheless, after the burn-in period is concluded, it will proceed to repeatedly select $a^-$, hence incurring arbitrary losses. If $a^+$ represented the act of shutting down, then this would mean that quite apart from detecting model mis-specification, this model would allow a harmful agent to permanently ignore instructions to shut down.

\citet{soares:2015:corrigibility} seem to want a shutdown button that works as a mechanism of last resort, to shut an AI system down in cases when it has observed and refused a programmer suggestion (and the programmers believe that the system is malfunctioning). Clearly, \emph{some} part of the system must be working correctly in order for us to expect the shutdown button to work at all. However, it seems undesirable for the working of the button to depend on there being zero critical errors in the specification of the system's prior, the specification of the reward function, the way it categorizes different types of actions, and so on. Instead, it is desirable to develop a shutdown module that is small and simple, with code that could ideally be rigorously verified, and which ideally works to shut the system down even in the event of large programmer errors in the specification of the rest of the system. In order to do this in a value learning framework, we require a value learning system that (i) is capable of having its actions overridden by a small verified module that watches for shutdown commands; (ii) has no incentive to remove, damage, or ignore the shutdown module; and (iii) has some small incentive to keep its shutdown module around; even under a broad range of cases where $R$, the prior, the set of available actions, etc. are misspecified.

It seems quite feasible to us that systems that meet the above desiderata could be described in a CIRL framework.

\section{Acknowledgements}
Thanks to Nate Soares and Matt Graves for feedback on draft versions.

\printbibliography

\end{document}